\documentclass[runningheads,utf8]{llncs}

\usepackage[utf8]{inputenc}
\usepackage[T1]{fontenc}
\usepackage{amsmath,amssymb,mathtools}
\usepackage{tikz}
\usepackage{pgfplots}
\usepackage{pgfmath}
\usepackage{expl3}
\usepackage{xstring}
\usepackage{xparse}
\usepackage{etoolbox}
\usepackage{xcolor}
\usepackage{graphicx}
\usepackage{ifthen}
\usepackage{csquotes}
\usepackage[pdftex, pdfborder={0 0 0}]{hyperref}
\usepackage{subcaption}
\usepackage{xspace}
\usepackage[textsize=small]{todonotes}
\usepackage[shortcuts]{extdash}
\usepackage{soul}
\usepackage{marvosym}
\usepackage{mathtools}

\newenvironment{itemize*}{\vspace*{-.5em}\begin{itemize}\setlength{\itemsep}{0pt}}{\end{itemize}\vspace*{-.5em}}

\newcommand{\ignoreLiteral}[1]{}

\newcommand{\sID}[2][]{\ensuremath{\textrm{ID}#1\ifstrempty{#2}{}{_{\textrm{#2}}}}\xspace}
\newcommand{\ID}{\sID{}}
\newcommand{\sGED}{\sID{GED}}
\newcommand{\sMLE}{\sID{MLE}}
\newcommand{\sMOM}{\sID{M\kern-.5ptO\kern-.5ptM}}
\newcommand{\sALID}{\sID{A\kern-.5ptL\kern-.6ptI\kern-.5ptD}}
\newcommand{\sTLE}{\sID{TLE}}
\newcommand{\sGAbider}{\sID{A\kern-1ptB\kern-.5ptI\kern-.5ptD}}
\newcommand{\sSAbider}{\sID{R\kern-.5ptA\kern-1ptB\kern-.5ptI\kern-.5ptD}}

\newcommand{\Cov}{\mathrm{Cov}}
\newcommand{\Jac}{\nabla}%

\pgfplotsset{compat=1.8}

\newcounter{tmparraycounter}
\newcommand{\arrayTmpSaveItem}[1]{%
    \stepcounter{tmparraycounter}%
    \expandafter\def\csname tmparrayname\thetmparraycounter\endcsname{#1}%
}
\newcommand{\arrayParse}[1]{%
    \setcounter{tmparraycounter}{0}%
    \renewcommand{\do}{\arrayTmpSaveItem}%
    \docsvlist{#1}%
}
\newcommand{\arrayElemNth}[2]{\arrayParse{#1}\csname tmparrayname#2\endcsname}

\def\arrayElemFirst#1{10}
\def\arrayElemLast#1{300}

\newcommand{\histAreaOpacity}{.5}
\newcommand{\histLineWidth}{1pt}
\newcommand{\histLegendScale}{.77}
\newcommand{\histLegendCols}{2}

\definecolor{orangeScheme1}{RGB}{242,200, 91}
\definecolor{orangeScheme2}{RGB}{251,164,101}
\definecolor{orangeScheme3}{RGB}{248,110, 81}
\definecolor{orangeScheme4}{RGB}{238, 62, 56}
\definecolor{orangeScheme5}{RGB}{209, 25, 62}

\definecolor{blueScheme1}{RGB}{ 83,204,236}
\definecolor{blueScheme2}{RGB}{ 25,116,211}
\definecolor{blueScheme3}{RGB}{  0,  1,129}

\definecolor{greenScheme1}{RGB}{204,255,204}
\definecolor{greenScheme2}{RGB}{179,230,185}
\definecolor{greenScheme3}{RGB}{153,204,166}
\definecolor{greenScheme4}{RGB}{128,179,147}
\definecolor{greenScheme5}{RGB}{102,153,128}
\definecolor{greenScheme6}{RGB}{ 77,128,108}
\definecolor{greenScheme7}{RGB}{ 51,102, 89}
\definecolor{greenScheme8}{RGB}{ 26, 77, 70}
\definecolor{greenScheme9}{RGB}{  0, 51, 51}

\definecolor{tolPalette1}{RGB}{ 51, 34,136}
\definecolor{tolPalette2}{RGB}{ 17,119, 51}
\definecolor{tolPalette3}{RGB}{ 68,170,153}
\definecolor{tolPalette4}{RGB}{136,204,238}
\definecolor{tolPalette5}{RGB}{221,204,119}
\definecolor{tolPalette6}{RGB}{204,102,119}
\definecolor{tolPalette7}{RGB}{170, 68,153}
\definecolor{tolPalette8}{RGB}{136, 34, 85}

\tikzset{
    histBarStyle/.style = {
        draw=#1,
        fill=#1,
        line width=\histLineWidth
    }
}

\definecolor{showcasePal1}{RGB}{ 51, 34,136}
\definecolor{showcasePal2}{RGB}{115,199,185}
\definecolor{showcasePal3}{RGB}{204,102,119}

\pgfplotscreateplotcyclelist{showcaseCycleList}{%
    {histBarStyle=showcasePal1},
    {histBarStyle=orangeScheme1},
    {histBarStyle=blueScheme1},
}

\pgfplotscreateplotcyclelist{estsCycleList}{%
    {histBarStyle=orangeScheme1},
    {histBarStyle=orangeScheme2},
    {histBarStyle=orangeScheme3},
    {histBarStyle=orangeScheme4},
    {histBarStyle=orangeScheme5},
    {opacity=0},
    {histBarStyle=blueScheme1},
    {histBarStyle=blueScheme2}
}

\pgfplotscreateplotcyclelist{nestedCycleList}{%
    {histBarStyle=tolPalette1},
    {histBarStyle=tolPalette2},
    {histBarStyle=tolPalette3},
    {histBarStyle=tolPalette4},
    {histBarStyle=tolPalette5},
    {histBarStyle=tolPalette6},
    {histBarStyle=tolPalette7},
    {histBarStyle=tolPalette8}
}

\ExplSyntaxOn
\newcommand{\histName}[1]{%
    \IfStrEqCase{#1}{%
        {GEDEstimatorFix}{\sGED}%
        {HillEstimator}{\sMLE}%
        {MOMEstimatorFix}{\sMOM}%
        {ALIDEstimatorFix}{\sALID}%
        {TightLIDEstimatorFix}{\sTLE}%
        {GeomAbider}{\sGAbider}%
        {StrictAbider}{\sSAbider}%
    }%
}
\ExplSyntaxOff

\newenvironment{histAreaAxis}[7][]{%
    \begin{axis}[%
        ybar,%
        xmin=#2,%
        xmax=#3,%
        ymin=0,%
        bar width=#4,%
        ymajorgrids,%
        scaled ticks=false,%
        y tick label style={%
            /pgf/number format/.cd,%
                fixed,%
                fixed zerofill,%
                precision=0,%
            /tikz/.cd%
        },%
        yticklabel={\pgfmathparse{\tick*100}\pgfmathprintnumber{\pgfmathresult}},%
        xtick pos=left,%
        ytick pos=left,%
        xlabel={\ID estimate},%
        ylabel={\% of points},%
        width=#6,%
        height=#7,%
        const plot mark mid,%
        every axis plot/.append style={%
            fill opacity=\histAreaOpacity,%
            draw opacity=0%
        },%
        cycle list name=#5,%
        #1%
    ]%
    \renewcommand{\addlegendentry}[1]{}%
}{%
    \end{axis}%
}
\newenvironment{histLineAxis}[7][]{%
    \begin{axis}[%
        xmin=#2,%
        xmax=#3,%
        ymin=0,%
        width=#6,%
        height=#7,%
        scaled ticks=false,%
        yticklabels={,,},%
        xticklabels={,,},%
        xtick pos=left,%
        ytick pos=left,%
        const plot mark mid,%
        every axis plot/.append style={%
            fill opacity=0%
        },%
        cycle list name=#5,%
        legend image post style={%
            fill opacity=\histAreaOpacity,%
            scale=\histLegendScale%
        },%
        legend style={%
            area legend,%
            at={(0,1)},%
            anchor=south west,%
            nodes={scale=\histLegendScale}%
        },%
        legend columns=\histLegendCols,%
        #1%
    ]%
}{%
    \end{axis}%
}

\definecolor{pcpGradCol1}{RGB}{211,148,  0}
\definecolor{pcpGradCol2}{RGB}{148,  0,211}
\definecolor{pcpGradCol3}{RGB}{  0,211,148}
\tikzset{
    pcpGradientA/.style={
        draw=pcpGradCol2!#1!pcpGradCol1
    },
    pcpGradientB/.style={
        draw=pcpGradCol3!#1!pcpGradCol2
    }
}

\newcommand{\lastplot}[3][]{}

\pgfplotsset{%
    every axis/.append style={%
        label style={font=\scriptsize},%
        tick label style={font=\tiny},%
        x tick label style={yshift=.25em},%
        y tick label style={xshift=.1em},%
        ylabel shift=-5pt,%
        xlabel shift=-5pt,%
        legend style={%
            draw=none,%
            fill=none,%
            font=\footnotesize,%
            shift={(0pt,2pt)}%
        },%
        title style={%
            font=\scriptsize,%
            shift={(0pt,-6pt)}%
        }%
    }%
}

\title{MESS: Manifold Embedding Motivated\\ Super Sampling}
\titlerunning{MESS: Manifold Embedding Motivated Super Sampling}
\author{Erik Thordsen \Letter\orcidID{0000-0003-1639-3534}
\and Erich Schubert\orcidID{0000-0001-9143-4880}} %
\authorrunning{E. Thordsen and E. Schubert}
\institute{
TU Dortmund University, Dortmund, Germany\\
\email{\{erik.thordsen,erich.schubert\}@tu-dortmund.de}
}

\begin{document}
\maketitle
\begin{abstract}
Many approaches in the field of machine learning and data analysis rely on the assumption that the observed data lies on lower-dimensional manifolds.
This assumption has been verified empirically for many real data sets.
To make use of this manifold assumption one generally requires the manifold to be locally sampled to a certain density such that features of the manifold can be observed.
However, for increasing intrinsic dimensionality of a data set the required data density introduces the need for very large data sets, resulting in one of the many faces of the curse of dimensionality.
To combat the increased requirement for local data density we propose a framework to generate virtual data points that faithful to an approximate embedding function underlying the manifold observable in the data.
\end{abstract}

\section{Introduction}\label{sec:introduction}
It is generally assumed that data is not entirely random but rather obeys some internal laws that can be considered a generative mechanism.
Whenever each sample is a numerical vector, one key observation is almost self-evident:
Correlating dimensions hint at possible causalities and are thereby useful for understanding the underlying generative mechanism.
In the most basic approach, one can apply principal component analysis to find the strongest correlations throughout the entire dataset and use, e.g., the five largest components as a simplifying model.
Semantically this corresponds to a generative mechanism that takes a five-dimensional vector and embeds it alongside the principal components in whatever many features are observed.
Therein already lie the limitations of this idea:
We can only represent linear embedding functions in this way.

Considering the case of a sine wave on a plot in two dimensions, the observed data is using only one parameter.
It would hence be reasonable to use a model that only uses a single parameter as well.
As a simple relaxation, we can assume the data to lie on any manifold.
This generalizes the concept of our generative mechanism to any \emph{locally} linear function.
However, this introduces a novel problem:
If we wish to use the methods developed for linear functions on locally linear functions, we introduce an error that grows with the observed locality.
In practice, that means if we analyze the manifold surrounding a single point, we can not take as many of the surrounding points as possible but must restrict ourselves to all points within some maximum radius.
In the field of intrinsic dimensionality (ID) estimation that focuses on estimating the number of parameters of one such embedding function, it is therefore common to use the $k$-nearest-neighbors for estimating the ID \cite{DBLP:conf/kdd/AmsalegCFGHKN15,DBLP:conf/sdm/AmsalegCHKRT19,tr/nii/ChellyHK16,DBLP:conf/icdm/HouleKN12,DBLP:conf/sisap/ThordsenS20}.
Yet, ID estimators require a sufficient amount of points in a neighborhood to be stable, which contradicts the requirement to observe as small a neighborhood as possible.
Ideally, we would like to use infinitesimal small neighborhoods containing infinitely many neighbors as in this case ID estimation can be investigated purely analytically \cite{DBLP:conf/sisap/Houle17,DBLP:conf/sisap/Houle17a,DBLP:conf/sisap/Houle20}.
Yet, as this is practically impossible, ID estimation is inherently approximate.

In this paper, we introduce a novel framework to generate additional data points from the existing ones that are meant to lie on, or very close to, the manifold of the generating mechanism.
We intend to increase the local point density on the manifold, thus decreasing the radius required for stable ID estimation.
Further applications, for example in machine learning, which could benefit from additional training data faithful to the generative mechanism, are outside the focus of this paper, yet certainly of interest for further research.

The remainder of this paper is structured as follows: Section~\ref{sec:related} surveys related work and the ID estimation approaches used in this paper.
The novel supersampling framework with its underlying theory is described in Section~\ref{sec:theory}.
The impact of the supersampling on ID estimators under varying parameterization is highlighted in Section~\ref{sec:experiments}.
The closing Section~\ref{sec:conclusion} gives an overview of the current state of the method and provides an outlook on future work.

\section{Related work}\label{sec:related}
In the past, the concept of intrinsic dimensionality has been analyzed within two major categories.
In the first category lie the geometrically motivated ID estimators like the PCA estimator, its local variant the lPCA estimator, the FCI estimator \cite{Erba2019IntrinsicDE}, or the recent ABID estimator \cite{DBLP:conf/sisap/ThordsenS20}, which uses the pairwise angles of neighbor points.
These estimators use measures like the local covariance of (normalized) points to estimate the spectrum of the neighborhoods, from which they deduce an estimate of the ID that mostly ignores weaker components as observational noise.
The second category of analytically motivated ID estimators contains, e.g., the Hill estimator \cite{doi:10.1214/aos/1176343247}, the Generalized Expansion Dimension estimator \cite{DBLP:conf/icdm/HouleKN12}, or more recent variants like the ALID~\cite{tr/nii/ChellyHK16} and TLE estimator~\cite{DBLP:conf/sdm/AmsalegCHKRT19}.
These are based on the idea that if the data lies uniformly in the parameter space, i.e.{} the preimage of the data under the embedding function, then distances in the data should increase exponentially in the ID at any point on the manifold.
These estimators can be considered the discrete continuation of analytical approaches like the Hausdorff dimension.
However, they are vulnerable to varying sampling densities.
As this work focuses on the improvement of estimates by supersampling the data and not on the comparative performance between estimators, we will use only a selection of estimators as prototypes for their categories.
The supersampling framework itself rests somewhere between these two categories:
It is generally motivated geometrically but the supersampled data can be understood as uncertainty added to the data set whose influence decreases over distance.
The framework can thus introduce information orthogonal to the ID estimation approaches and potentially enrich estimators from both categories.

The concept of enriching data sets by introducing new points is by itself not a new idea.
Domain-specific supersampling techniques are very common for example in image-based machine learning, where images are created by adding noise, scratches, color or brightness shifts, flipping or rotating the image, and many more.
The first few can be thought of as adding noise to (some of) the features while the latter correspond to geometric operations like rotation or translation.
These operations, although possibly faithful to the manifold of the generative mechanism (e.g., amount of noise in the imaging hardware), are not necessarily faithful to the manifold presented in the analyzed data set.

There are, however, approaches that in a very general manner attempt to mimic the data with additional points.
One of these is the highly-cited SMOTE approach introduced by Chawla et al.~\cite{Chawla2002SMOTESM} which introduces new points by adding a series of linear interpolations of points in the feature space.
On locally non-convex manifolds this approach creates points that do not lie on the manifold.
For machine learning tasks, this deviation is likely tolerable as, e.g., barely overlapping classes are still separated.
In terms of ID estimation, however, even small systematic deviations from the manifold can introduce new orthogonal components resulting in overestimation.
Later variants of the SMOTE approach have tried to work around this issue but focus on the machine learning objective of not mixing classes.

The objectives of supersampling (or oversampling in the words of SMOTE) for machine learning and ID estimation are quite similar yet inherently different.
For machine learning, a more redundant sampling of the manifold is useful for parameter optimization yet possibly diminishes the model robustness compared to data scattered around the manifold.
In ID estimation, we need an exact representation of the manifold without additional orthogonal components.
One can therefore consider the constraint set of supersampling for ID estimation as stronger than that for machine learning.
Solutions for ID estimation are applicable for machine learning but not vice versa.

The framework by Bellinger et al.~\cite{DBLP:journals/ml/BellingerDJ18} is similar to the framework introduced here, as it also attempts to take additional samples from the manifold itself.
They, however, base their method on manifold learning approaches, which to give decent results, already require quite dense data on the manifold and additionally a parameterization of the intrinsic dimensionality.
Starting with a method that requires ID estimates to improve ID estimates is a circular argument that makes the framework of Bellinger et al.~inapplicable for ID estimation.
This problem also affects manifold-based variants of SMOTE, e.g., the one by Wang et al.~\cite{wang2006classification} which involves locally linear embeddings.

At last, multiple techniques in computer graphics have been studied to improve, e.g., the resolution of point cloud scans of objects.
These approaches, however, mostly focus on improving 2D surfaces in 3D space and do not generalize to arbitrary dimensions of both parameter and feature space, partly as information like surface normals used in these methods do not generalize either.

\section{Manifold faithful supersampling}\label{sec:theory}
As mentioned in Section \ref{sec:related}, the framework for supersampling introduced in this paper is based on geometric observations and aims to create samples from the manifold that is created by the generative mechanism via an embedding function from some parameter space.
It hence shares the manifold assumption implied by many downstream applications like autoencoders and ID estimators like ABID~\cite{DBLP:conf/sisap/ThordsenS20} or ALID \cite{tr/nii/ChellyHK16}.
Whilst Chawla et al.~\cite{Chawla2002SMOTESM} and Bellinger et al.~\cite{DBLP:journals/ml/BellingerDJ18} named their methods over-sampling, it will here be called supersampling in line with the computer graphics semantics:
We increase the amount of data to reduce undesired artifacts and improve the \enquote{bigger picture} of the original data set.

The remainder of this section will consist of the geometric observations underlying the framework, an explanation of the generation pipeline, followed by different options for the modules that make up the pipeline.

\subsection{Approaching the embedding function}
We assume that the data lies on some manifold.
To be more precise, we assume that our data $X$ obeys some locally linear%
\footnote{
	The aspect of local linearity is disputable, as it does not allow for singularities and is weak on functions with high curvature.
	However, the typically low sampling density of real data sets makes geometric methods ineffective on non-locally-linear functions.
}
embedding function $E$ such that for every $x \in X$ we have some preimage $\tilde{x}$ such that $E(\tilde{x}) = x$.
Yet data is often noisy, which is why this will likely not hold for all $x \in X$, and for the rest, we probably only have $E(\tilde{x}) \approx x$.
While we can tolerate that the embedding function is only approximate, we will have to accept that we can not know which $x \in X$ are noise points and are unrelated to the embedding function.
It is, therefore, reasonable to assume that this assumption holds for all $x$, as we cannot correct it for the other points either way, except for pre-filtering $X$ with some outlier detection method.
In the remainder of this paper, the tilde over a data point or data set will represent the preimage as per the assumed embedding function.
We denote the ground truth ID as $\delta$ and the observed feature dimension as $d$.

Assuming $E: \mathbb{R}^\delta \rightarrow \mathbb{R}^d$ to be locally linear, in the infinitesimal we obtain %
\begin{equation}
	\Cov[E(B_\varepsilon(\tilde{x}))]
	\xrightarrow[\varepsilon \rightarrow 0^+]{}
	\frac{\varepsilon^2}{\delta+2} \Jac E(\tilde{x})\Jac E(\tilde{x})^T
\end{equation}
where $\Cov[E(B_\varepsilon(\tilde{x}))] \in \mathbb{R}^{d \times d}$ is the covariance matrix of all points from an $\varepsilon$-ball around $\tilde{x}$ embedded with the embedding function $E$, $\Jac E(\tilde{x}) \in \mathbb{R}^{d \times \delta}$ is the Jacobian of $E$ evaluated at $\tilde{x}$.
This formula can be easily derived from the propagation of uncertainty \cite{sengupta2003linear} taking the Jacobian as the linear approximation of $E$.
The covariance matrix of the $\delta$-ball with radius $\varepsilon$ can therein be substituted by $\tfrac{\varepsilon^2}{\delta+2}$ as it is a diagonal matrix with $\tfrac{\varepsilon^2}{\delta+2}$ on the diagonal \cite{Joarder2008OnTD}.

Naturally, we do not have such a high sampling density on real data sets that we can consider arbitrarily small neighborhoods.
We hence assume the simpler approximation
\begin{equation}
	\Cov[N_k(x)]
	\approx \Jac E(\tilde{x}) C \Jac E(\tilde{x})^T
\end{equation}
where $\Cov[N_k(x)]$ is the covariance matrix of the $k$-nearest-neighbors of $x$ in $X$, and $C$ is a diagonal matrix with scaling factors for each component gradient.
We here exchanged the infinitesimal neighborhood for the $k$-nearest-neighbors which introduces the approximation.
We also exchanged the scaling scalar for a scaling diagonal matrix, as curvature along the parameter geodesics and varying sampling density in parameter space can contract or expand the components of the Jacobian.
Solving for $\Jac E(\tilde{x})$ is futile though, as it is underdetermined from this approximation.
Yet, the rough approximation of aspects of $E$ via covariance matrices gives a glimpse at the manifold structure underlying the data set.

To stretch data according to some covariance matrix $\Sigma$, it is well known that the lower diagonal matrix $L$ of the Cholesky decomposition ($LL^T{=}\Sigma$) can be used.
Starting with a set of points $x_i$ sampled from $\mathcal{N}(0, 1)$ we can, hence, obtain points $x_i' = Lx_i$ which are distributed according to $\mathcal{N}(0,\Sigma)$.
That is, we can use the Cholesky decomposition to mimic the local linear approximation of the embedding function $\Jac E$.
Using that approximation of $\Jac E$ we can then compute locally linear embeddings of arbitrary distributions in parameter space.%

To compute the covariance matrices of the $k$-nearest-neighbors $N_k(x)$ of some point $x$, we use the biased formula $\tfrac{1}{k} N^TN$ where $N {\in} \mathbb{R}^{k {\times} d}$ contains the $k$-nearest-neighbors subtracted by $x$ as row vectors.
This virtually puts $x$ in the center of the distribution and is less susceptible to introducing additional orthogonal components due to manifold curvature.
Using the query point $x$ as the center of the distribution is common practice in ID estimation \cite{tr/nii/ChellyHK16,DBLP:conf/sdm/AmsalegCHKRT19,DBLP:conf/sisap/ThordsenS20}.
In addition, we add a small constant to the diagonal of all covariance matrices whenever computing any Cholesky or eigendecomposition to avoid numerical errors.

\subsection{Supersampling pipeline}
\begin{figure}[tb]\centering
	\def\lwidth{11em}
	\def\lheight{10em}
	\begin{subfigure}[t]{\lwidth}
		\centering
		{original data}
		\includegraphics[height=\lheight]{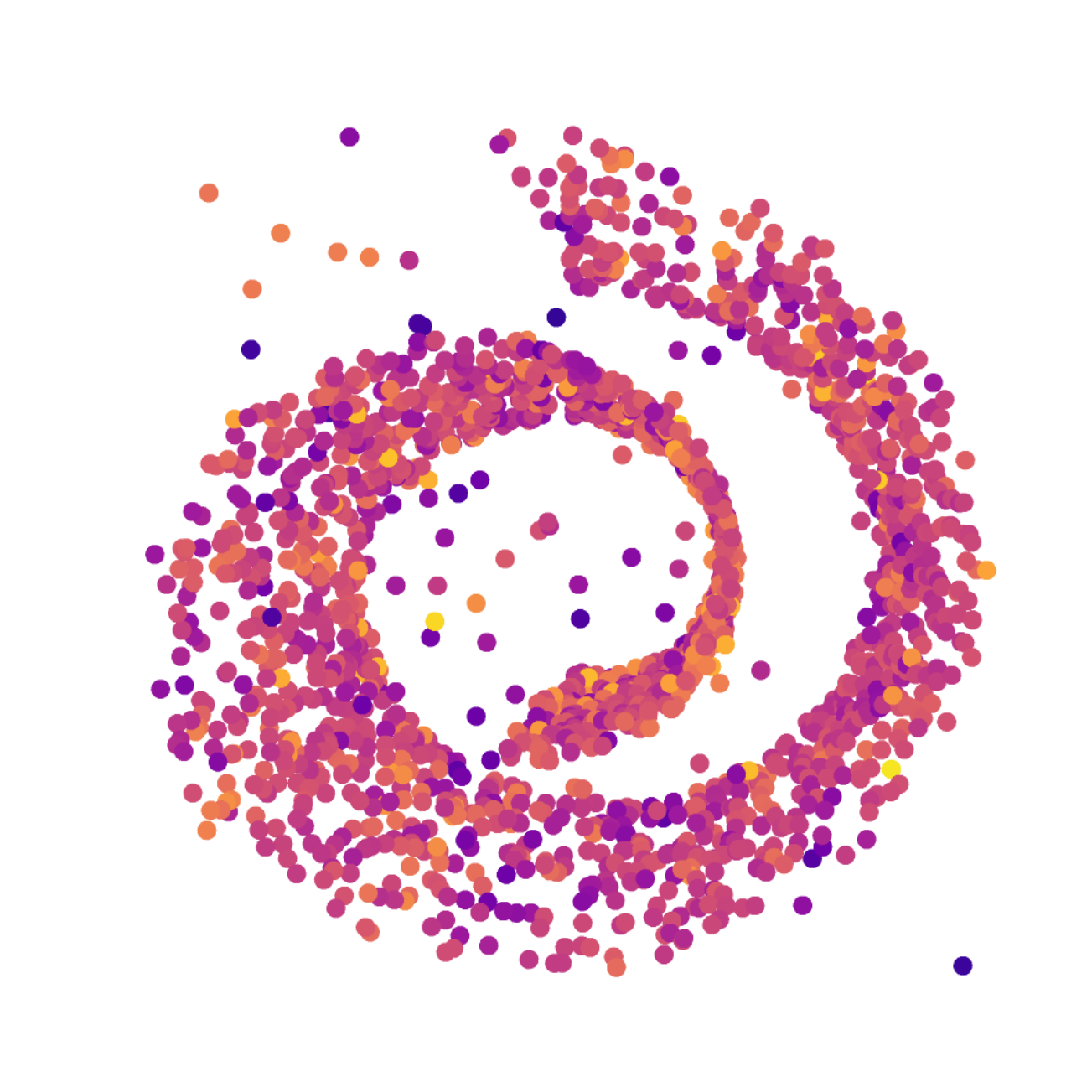}
	\end{subfigure}
	\begin{subfigure}[t]{\lwidth}
		\centering
		{without correction}
		\includegraphics[height=\lheight]{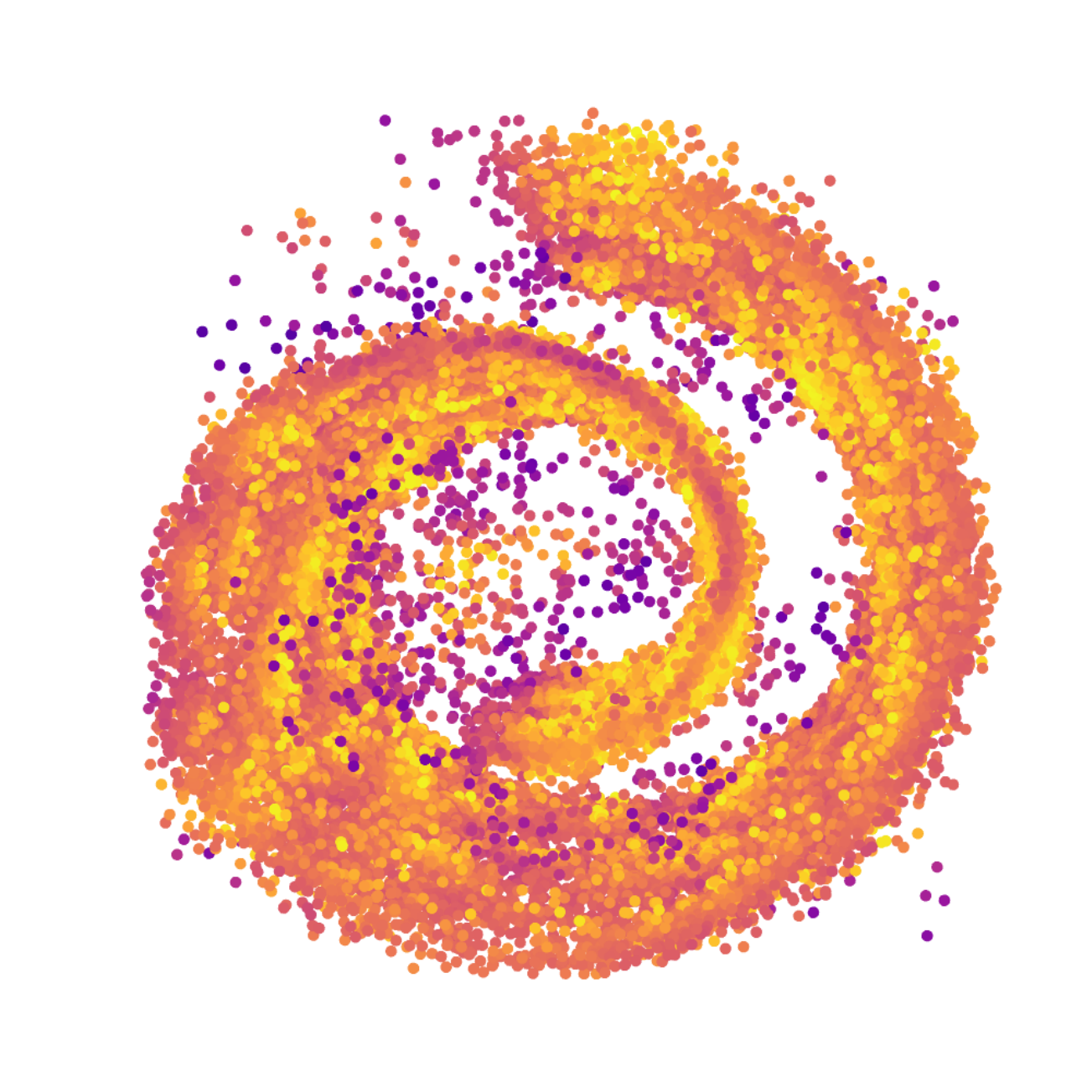}
	\end{subfigure}
	\begin{subfigure}[t]{12em}
		\centering
		{with correction}
		\includegraphics[height=\lheight]{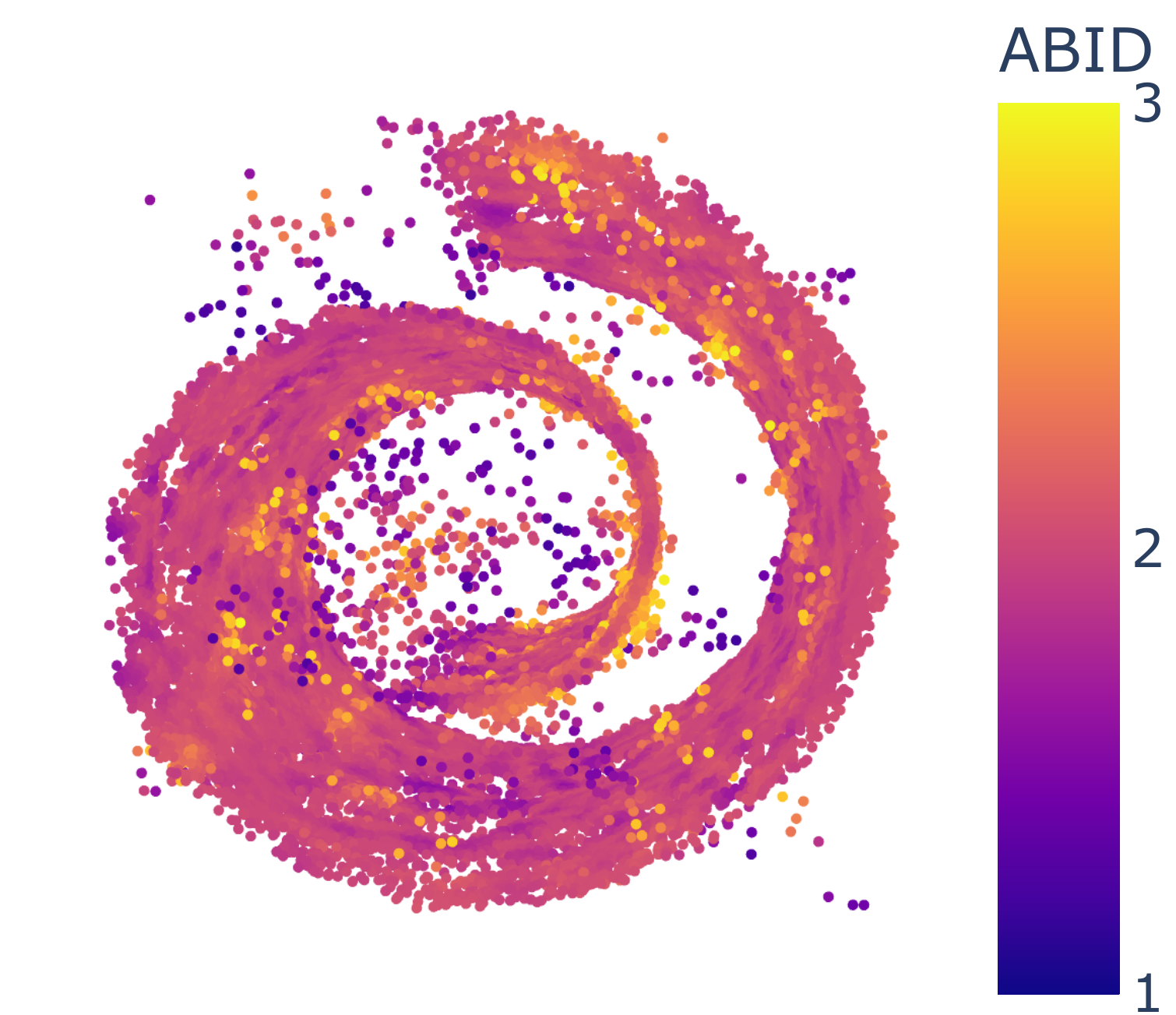}
	\end{subfigure}
	\caption{
		Supersampling a swiss roll with normal noise in feature space ($\sigma = 0.15$) and added uniform noise points.
		The left image shows the original data, the middle image is supersampled 50$\times$ using covariances and
		the right is the supersampled data \enquote{corrected} onto the manifold.
		All plots are colored by ABID estimates with 10 neighbors for the original data and 500 for the others.
	}
	\label{fig:pointcloud:swiss_roll}
\end{figure}

As explained in the previous section, we can use the covariance matrices of neighborhoods to describe local linear approximations of the embedding function.
From these local approximations, we can then sample some points that are already close to or even on the manifold.
However, contrary to Chawla et al.~\cite{Chawla2002SMOTESM} and Bellinger et al.~\cite{DBLP:journals/ml/BellingerDJ18} we will afterward modify each of the samples to better constrain them to the manifold.
Would we use the generated points without further processing, they would frequently lie outside the manifold whenever it has non-zero curvature, or when our observations are noisy.
This can be seen in Fig.~\ref{fig:pointcloud:swiss_roll} as the surface of the corrected supersampling is much smoother than the raw supersampling with less spread orthogonal to the swiss roll.
To move the supersampled points onto the manifold, we use weighted means of candidate points that are more aligned with the covariances of the nearby original data points.
For ID estimation we then chose the $k$-nearest-neighbors from the supersampled points rather than the original data set.
The $k$ used for the ID estimates can then be larger by a factor up to the number of samples per original point without increasing the neighborhood radius.
We, hence, start with some data set $X$, supersample it by a factor of $\mathtt{ext}$ to $X_{ext}$, move the supersamples onto the manifold obtaining $X_{corr}$ and then compute ID estimates for each $x \in X$ using the $(\mathtt{ext} \cdot k)$-nn of $x$ in $X_{corr}$.

\subsection{Supersampling modules}
The approach in this general form allows for many variants of each step.
We can, for example, generate initial samples from a multivariate normal distribution around the original samples using the locally estimated covariance matrix.
The resulting points would certainly lie close to, or on, the local linear approximation of the manifold.
To structure this section, we will first describe the proposed generation modules, followed by the correction modules.

\smallskip\noindent
\textbf{Sample generation:\enskip}
The generated samples can have two kinds of \enquote{closeness} to the manifold:
Either close to the original points in feature space, or try close in the hypothesized parameter space.

The first type of generation rules generates samples close to the original points in feature space.
Samples can be drawn from normal distributions or $d$-balls around our original points.
For the radius of the $d$-ball or the standard deviation of the normal distribution, we can, e.g., use distances to the $k$-th nearest neighbor.
These approaches, however, generally introduce additional orthogonal components as they do not obey the shape of the manifold and in initial experiments performed quite poorly.
We, therefore, did not further analyze these generation rules.

The second type of generation rules mimics the local linear approximation of the manifold, thereby generating points close in the hypothesized parameter space rather than feature space.
The first rule of this type that we propose uses the covariance-based multivariate normal distribution.
Points generated by this rule can be expected to closely follow the local linear approximation of the manifold, yet have an increasing density towards the original points.
The resulting set of samples, hence, is all but uniformly distributed in parameter space and can thereby potentially encumber expansion rate-based ID estimators.

To compensate for the non-uniform density of multivariate normal distributions, we propose another rule which is based on the Cholesky decomposition of the covariance matrix ($L$ with $LL^T {=} \Sigma$).
The Cholesky decomposition is easy to compute and gives a linear map that maps a unit $d$-ball with a Mahalanobis distance of $\leq 1$ with respect to $\Sigma$.
Using this map, we can then transform uniform samples from a unit $d$-ball into samples within one standard deviation according to $\Sigma$.
Assuming the Cholesky decomposition to be a linear approximation of the embedding function, this produces points uniformly at random in parameter space whenever $\delta \approx d$.
Yet ID estimation is generally performed on data sets where the ID is much lower than the number of features ($\delta \ll d$), where the points would again be very concentrated around the original points.

The problem therein is simple:
To generate points at the exact expansion rate of the parameter space, one needs to know the dimension of the parameter space.
We can, in general, not know the true ID, yet an initial estimate could be close enough for generated data to be approximately uniform in parameter space.
But we cannot simply scale the lengths of samples uniform from the unit $d$-ball by exponentiating with $d/\delta$ as the lengths are not equally distributed along the $\delta {\ll} d$ components of~$\Jac E$.
To scale along the components of $\Jac E$ according to their lengths, we can use the eigenvalues of the Cholesky decomposition.
However, we can also skip the Cholesky decomposition and use the eigendecomposition of the covariance matrix ($V \Lambda V^T {=} \Sigma$), where $V \Lambda^{\frac{1}{2}}$ is assumed to be approximately~$\Jac E$.
Starting with samples uniform at random in a $d$-ball, we can multiply them with~$\Lambda^{\frac{1}{2}}$ and project them onto the sphere.
The resulting angular distribution is then compliant to that of a multivariate normal distribution using $\Lambda^{\frac{1}{2}}$ as the covariance matrix.
By scaling each of these points with~$r^{\delta'}$, where $r$ is uniform at random from 0~to~1 and $\delta'$ is the initial ID estimate we obtain a distribution that is compliant to the expansion of a $\delta'$-ball.
Finally multiplying with $\Lambda^{\frac{1}{2}}V^T$ (or $L$, though we already have the Eigen- but not the Cholesky decomposition) embeds the points according to~$\Jac E$.
Scaling with $\Lambda^{\frac{1}{2}}$ before normalization does not eliminate $d-\delta'$ components nor give equal weight to the remaining components, thereby not creating a uniform distribution in $\delta'$ components.
However, it is less reliant on knowing the exact $\delta$ and works for fractal $\delta'$ and is hence preferable in this early stage.
The resulting samples are approximately uniformly distributed in parameter space whenever $\delta \approx \delta'$.

As the regions in which samples are generated are overlapping, it is not obvious whether the increased effort from multivariate normal distributions to Eigendecomposition-based samples yields a clear improvement of sampling density in parameter space.
Besides that, not all ID estimators are inherently sensitive to misleading expansion rates.
In contrast, the more we enforce a proper uniform $\delta$-ball sampling in parameter space, the more we increase the average distance from our original points.
Improving on the sampling density might thus even result in more points that are further away from the manifold.
We, therefore, propose all three generation rules (covariance, Cholesky, $\delta$-ball) and provide a comparison in the evaluation section.

Due to the overlapping sampling regions, using an ID estimate for the sample generation is not the same circular argument mentioned in Section \ref{sec:related}.
Even if our ID estimate is quite different from the true ID, the generated data will still be more or less dense around the original points.
When using $(\mathtt{ext}\cdot k)$-nearest neighbors for estimation, the average expansion rate will remain similar to that of the original data.
We neither drop nor induce additional orthogonal components, which a badly parameterized manifold learning might do.
The ID estimates used for sample generation, therefore, have much less influence on the manifold shape and the final ID estimates than in the approach of Bellinger et al.~\cite{DBLP:journals/ml/BellingerDJ18}.

\smallskip\noindent
\textbf{Sample correction:\enskip}
Once we have created some samples for each of our original points, we could immediately compute ID estimates.
However, as displayed in Fig.~\ref{fig:pointcloud:swiss_roll}, the samples might be too noisy to describe the manifold.
Aside from the linear approximation diverging too strongly from the manifold in areas of high curvature, the covariance matrices in areas of high curvature contain non-zero orthogonal components to the manifold.
Thereby, the approximation of $\Jac E$ via the covariance matrix adds additional noise.
Original points that are not exactly on the manifold but have additional noise in the observed feature space introduce additional errors in our approximation.
It is, therefore, safe to assume, that on curved manifolds, or in the presence of high dimensional observational noise, supersamples frequently lie outside the manifold.
To constrain the generated samples back onto the manifold, we propose the following general approach:
For each supersample $p \in X_{ext}$ we search for its $k$-nn $x_1, \ldots, x_k \in X$ in the original data. We then generate candidates $c_i$ and weights $w_i$ dependent on $x_i$ and $p$ for each $1 \leq i \leq k$ and use the weighted mean over the $c_i$ as a corrected supersample.
By combining possible realizations of the maps $x,p \mapsto c$, and $x,p \mapsto w$, we obtain a set of correction rules.
As candidate maps, we propose: %
\begin{align*}
C_1: x,p &\mapsto \Sigma_x (p-x) \tfrac{\Vert p-x \Vert}{\Vert \Sigma_x (p-x) \Vert} + x \\
C_2: x,p &\mapsto L_x (p-x) \tfrac{\Vert p-x \Vert}{\Vert L_x (p-x) \Vert} + x
\end{align*}
where $\Sigma_x$ is the covariance matrix for $x$ and $L_x$ is its Cholesky decomposition.
The purpose of these maps is to rotate $p$ around $x$ towards a direction at which the probability density of the estimated multivariate normal is higher.
Each candidate is hence a more likely observation at a fixed distance for the $k$-nn.
The map $C_1$ multiplies each point with the covariance matrix, which equates to scaling the points along the principal components with a factor of the variance in these directions.
Components orthogonal to the manifold should have a comparably very small variance, which nearly eliminates all components weaker than those tangential to the manifold.

The second candidate map $C_2$ gives a similar result.
The Mahalanobis distance, which gives the distance in units of \enquote{standard deviations in that direction}, of the candidates prior to rescaling can be written as
\begin{align*}
	\sqrt{\rule[-2pt]{0pt}{10pt}\smash{(L_x(p-x))^T\Sigma_x^{-1}(L_x(p-x))}}
	&= \sqrt{\rule[-2pt]{0pt}{10pt}\smash{(p-x)^TL_x^T(\Sigma_x^T)^{-1}L_x(p-x)}}\\
	&= \sqrt{\rule[-2pt]{0pt}{10pt}\smash{(p-x)^TL_x^T(L_x^TL_x)^{-1}L_x(p-x)}}\\
	&= \sqrt{\rule[-2pt]{0pt}{10pt}\smash{(p-x)^T(p-x)}} = \Vert p-x \Vert.
\end{align*}
This candidate map, therefore, maps the Euclidean unit sphere onto the Mahalanobis unit sphere, effectively equalizing the influence of different components for $p$ to their relative strengths in $\Sigma_x$.

Where using the Cholesky decomposition neutralizes components orthogonal to the manifold approximated by $\Sigma_x$, the covariance matrix actively reinforces the components tangential to the manifold.
Both maps move samples onto the manifold.
$C_1$ increases the density along the larger components of $\Sigma_x$, whereas $C_2$ leaves the density about equal at the cost of being less strict.%

\begin{figure}[tb]\centering%
	\def\lwidth{12em}%
	\setlength{\fboxsep}{0pt}%
	\setlength{\fboxrule}{.2pt}%
	\begin{subfigure}[t]{\lwidth}
		\fbox{\includegraphics[width=\lwidth]{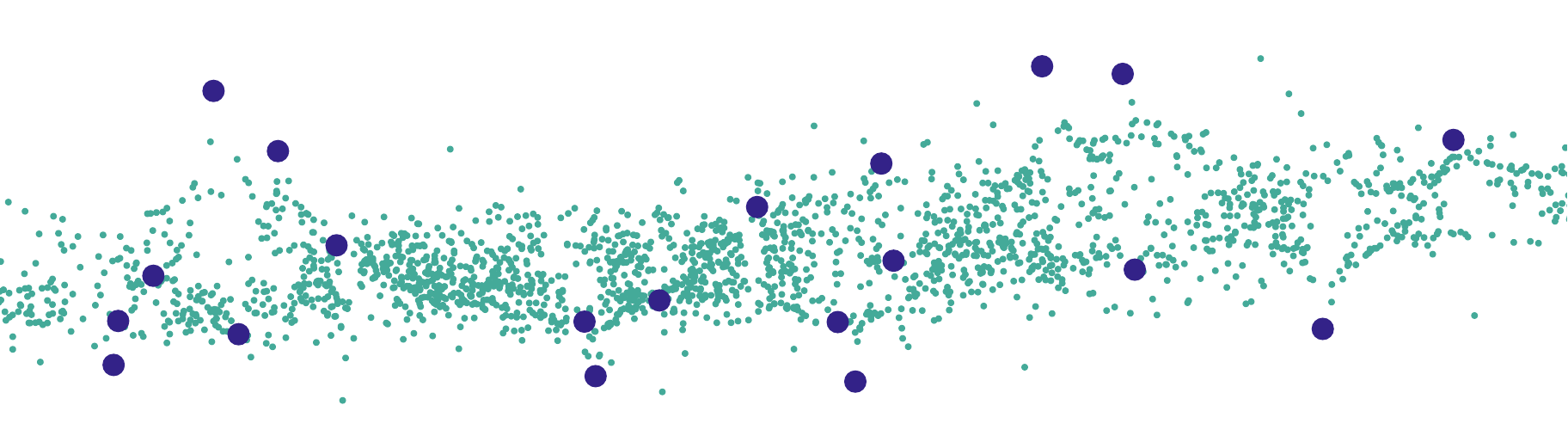}}\\[-.1em]
		\fbox{\includegraphics[width=\lwidth]{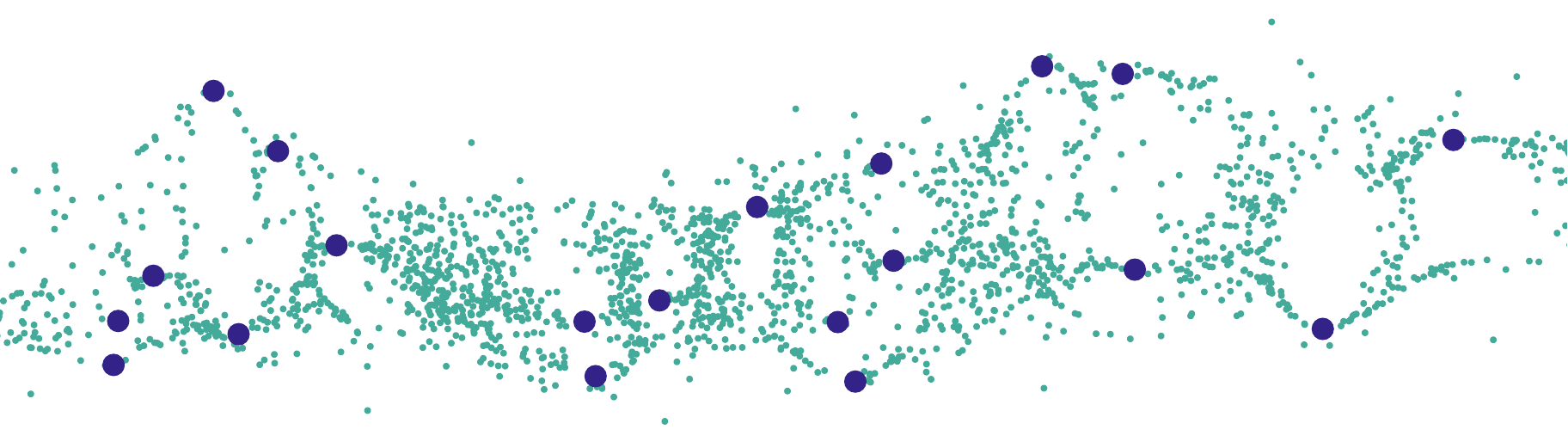}}
	\end{subfigure}
	\!
	\begin{subfigure}[t]{\lwidth}
		\fbox{\includegraphics[width=\lwidth]{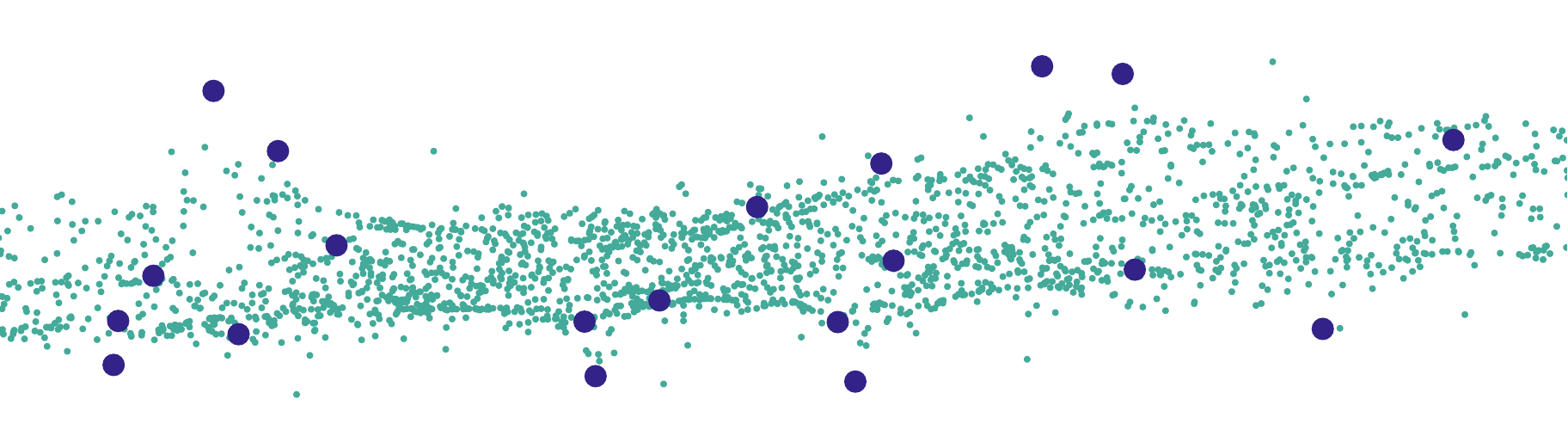}}\\[-.1em]
		\fbox{\includegraphics[width=\lwidth]{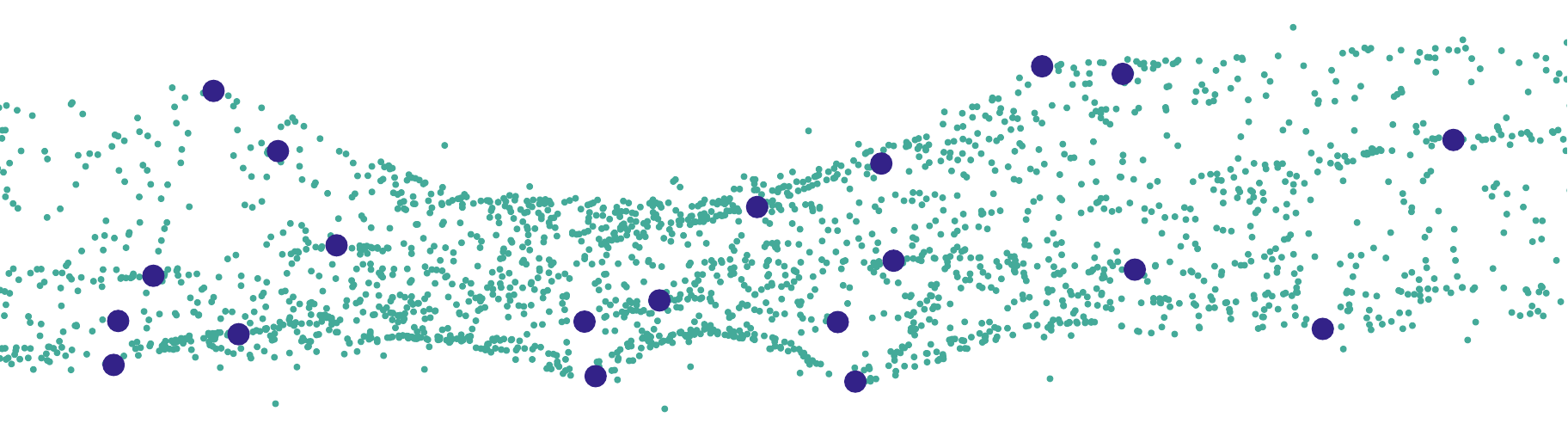}}
	\end{subfigure}
	\!
	\begin{subfigure}[t]{\lwidth}
		\fbox{\includegraphics[width=\lwidth]{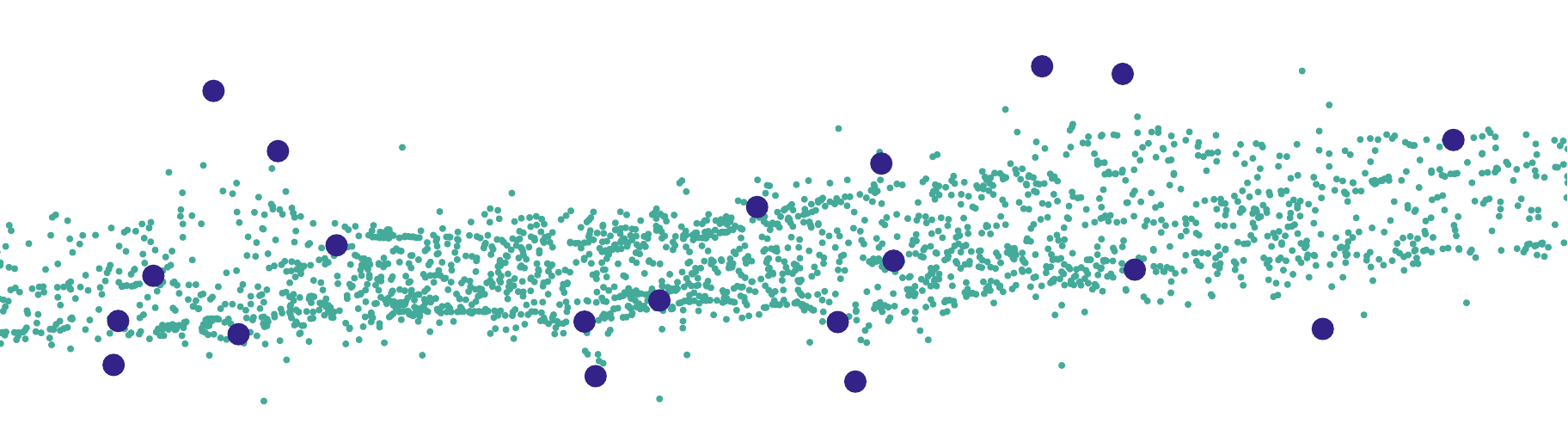}}\\[-.1em]
		\fbox{\includegraphics[width=\lwidth]{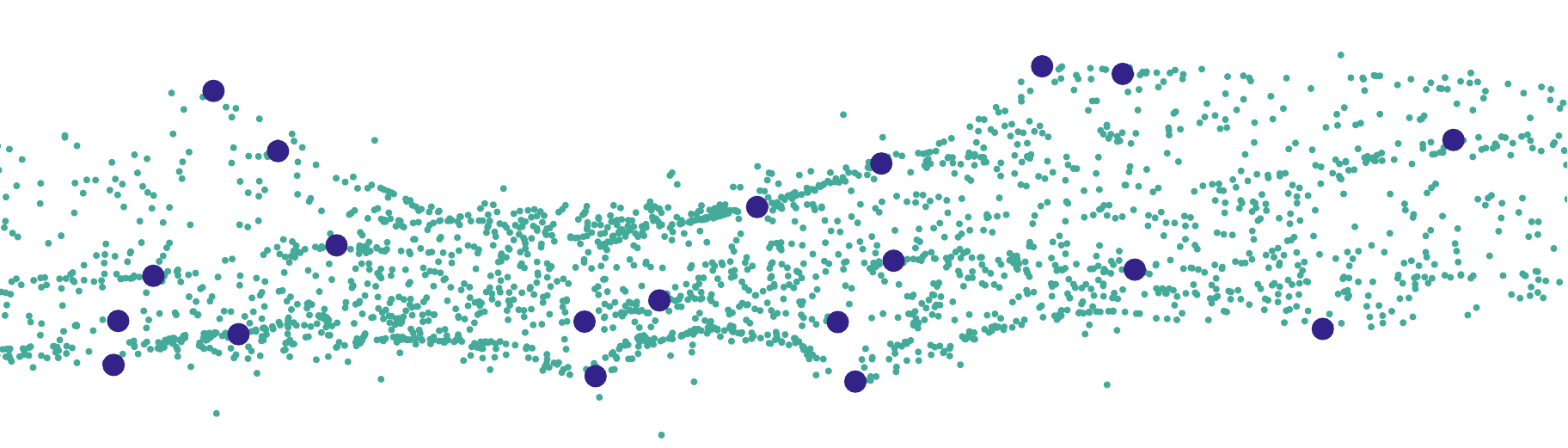}}
	\end{subfigure}
	\caption{
		Corrected supersampling of the dark original points for different weighting rules. From left to right are the weighting rules $W_1, W_2$, and $W_3$. The top row is without powers and the bottom row with a constant power of 2.
	}
	\label{fig:idw_scatters}
\end{figure}

For the weights, we propose an inverse distance weighted (IDW) scheme:%
\begin{align*}
	W_1: x,p &\mapsto \Vert p-x \Vert^{-1}\\
	W_2: x,p &\mapsto \sqrt{\rule[-2pt]{0pt}{10pt}\smash{(p-x)^T\Sigma_x^{-1}(p-x)}}^{-1}\\
	W_3: x,p &\mapsto \sqrt{\rule[-2pt]{0pt}{10pt}\smash{(x-p)^T\Sigma_p^{-1}(x-p)}}^{-1}
\end{align*}
Using IDW means enforces the corrected supersample set to be interpolating for the original data set, that is, if we sampled all possible points, the sample would pass through all of our original points.
This property is necessary to use the original points as centers for ID estimation as otherwise, they could be outside of the manifold spun by the supersampling.
In our experiments, we observed that $W_2$ and $W_3$ gave largely similar results, as the samples were already close to the manifold resulting in locally similar covariance matrices.
With $W_1$ the original points have a very strong pull on the samples, occasionally introducing dents in the manifold when the original data is noisy.
With $W_2$ and $W_3$, the corrected points give a smooth manifold approximation that is robust against noise on the original data set.
If the covariance matrices have a high variance among the $x_i$ and $p$, $W_3$ moves the samples into a slightly more compact shape than $W_2$ as single outlying $x_i$ have less impact on the correction.
These effects are displayed in Fig.~\ref{fig:idw_scatters}.
However, it vanishes for increasing neighborhood sizes.

In addition to using IDW means, we used powers of the weights to simulate the increase of the Mahalanobis sphere (points with equal Mahalanobis distance) surface area.
The exponents, as with the $\delta$-balls described for sample generation, should ideally be equal to the ID~$\delta$.
Fig.~\ref{fig:idw_scatters} displays the impact of both the different weighting rules and powers.
With powered weights, each original point has a much stronger pull on the nearby samples, reducing the tendency to interpolate larger trends.
The estimate quality seemed to rather decrease in initial experiments with powered weights even with ground truth ID values, which is why we did not further pursue this.

\section{Evaluation}\label{sec:experiments}

In the experiments, we applied the supersampling framework to the Hill estimator \cite{doi:10.1214/aos/1176343247} and the ABID estimator \cite{DBLP:conf/sisap/ThordsenS20} as prototypes for expansion- and geometry-based estimators.
We chose these two estimators specifically because they can be evaluated very quickly even on very large neighborhoods.
This is very important, as upscaling the data set size by a factor of, e.g., 200 also increases the neighborhoods used for ID estimation by up to the same scale.
If we consider 20 neighbors for ID estimation and supersampling covariances, the closest 4000 points in the supersampling should describe the same geometry as the initial 20 neighbors.
Estimators with super-linear complexity in the size of the neighborhood like the ALID \cite{tr/nii/ChellyHK16} or TLE estimators \cite{DBLP:conf/sdm/AmsalegCHKRT19} are slower when using supersampling.

As initial ID estimates, we used the ABID estimator \cite{DBLP:conf/sisap/ThordsenS20} as it is geometrically motivated like the MESS framework and should give a close enough ID estimate even for smaller neighborhoods on which the Hill estimator can give too large estimates especially for outlying points \cite{DBLP:conf/kdd/AmsalegCFGHKN15,tr/nii/ChellyHK16,DBLP:conf/sisap/HouleSZ18,DBLP:conf/sisap/ThordsenS20}.
The number of neighbors to compute the covariance matrices of the original points is called $\mathtt{k_1}$, the number of neighbors for the covariances of samples as well as the correction rules is called $\mathtt{k_2}$ and the number of samples to use for ID estimation of the original points is called $\mathtt{k_3}$.
The number of samples generated per original point is $\mathtt{ext}$.
Unless otherwise stated we used $\mathtt{k_2} := \mathtt{k_1}$ and $\mathtt{k_3} := \mathtt{k_1} \cdot \mathtt{ext}$.

\begin{figure}[tb]
	\centering
	\def\lheight{8em}
	\def\lhalfheight{4em}
	\begin{subfigure}{\lheight}
		\includegraphics[height=\lheight]{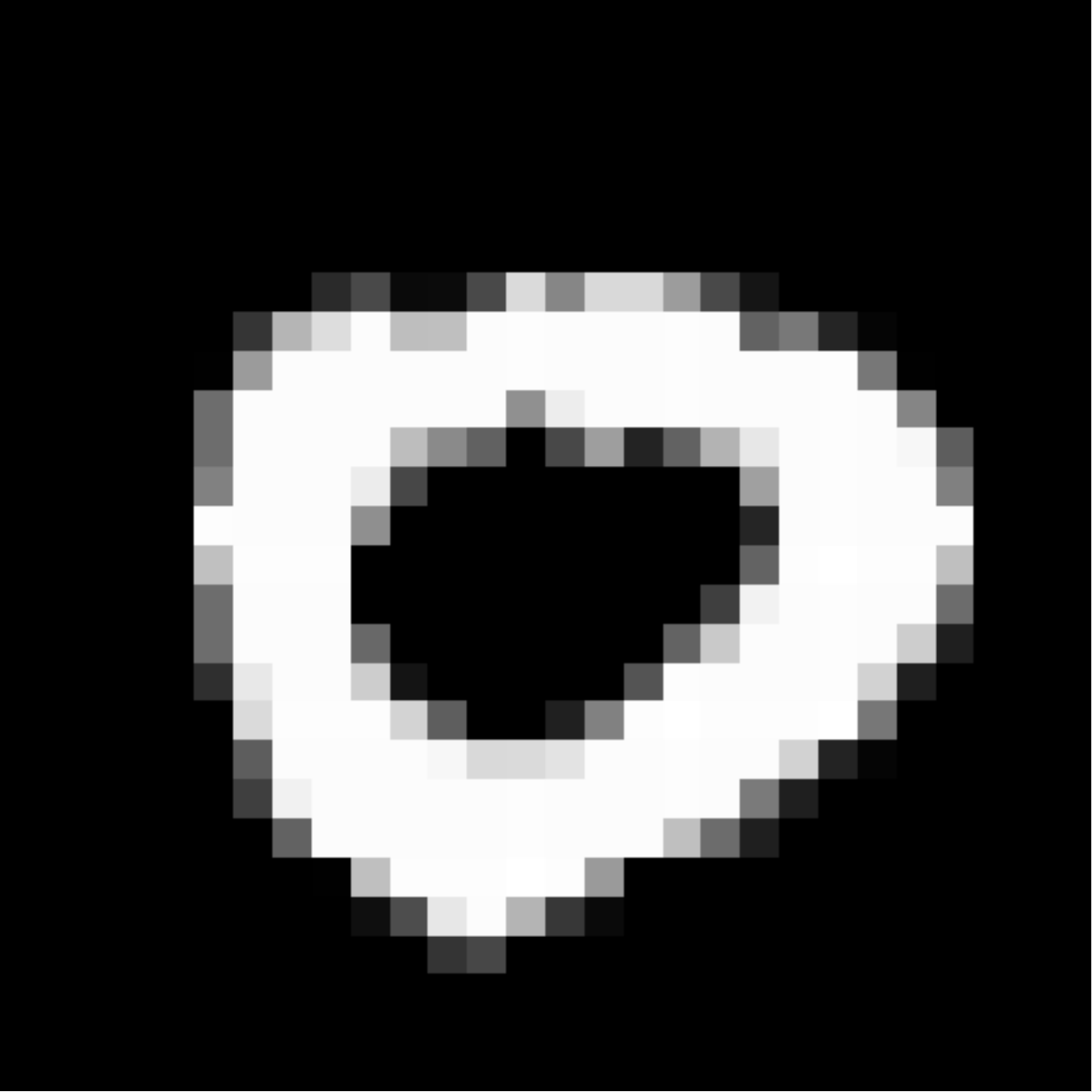}
	\end{subfigure}
	\begin{subfigure}{8em}
		\begin{tabular}{r}
			raw samples\\[.8em]
			nearest neighbors\\[.8em]
			corrected samples\\[.8em]
			nearest neighbors
		\end{tabular}
	\end{subfigure}
	\begin{subfigure}{20em}
		\includegraphics[height=\lhalfheight]{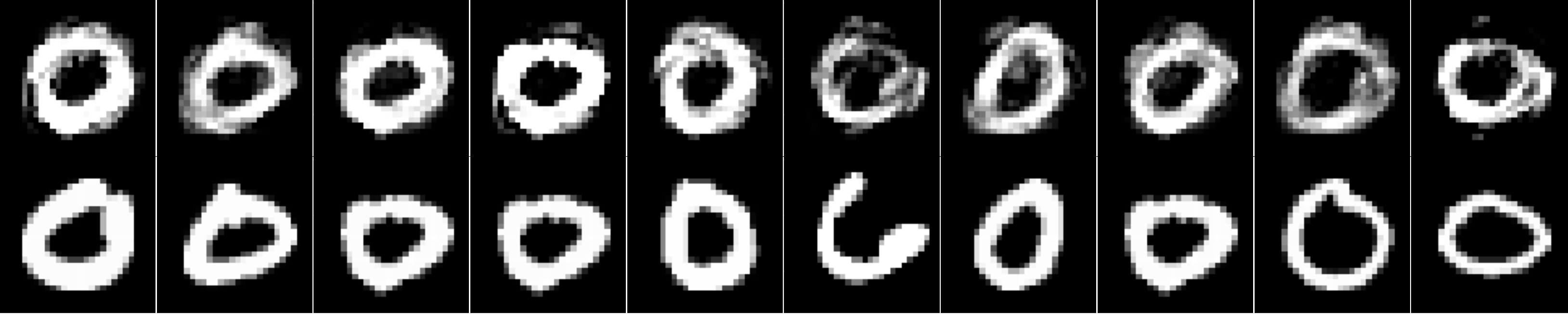}\\[-.1em]
		\includegraphics[height=\lhalfheight]{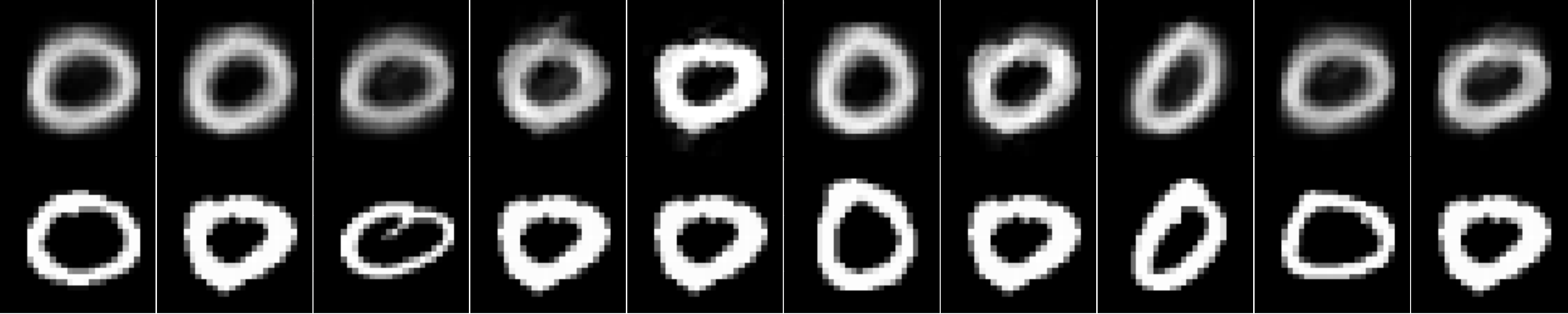}
	\end{subfigure}
	\caption{
		Supersamples around one of 1000 random MNIST images with $\mathtt{k_1}=20$.
		The original image is displayed on the left.
		The top two rows are raw covariance-based supersamples with their nearest neighbor in the original data set.
		The bottom two rows are the corrected ($C_1 + W_3$) supersamples with their nearest neighbor in the original data set.
		Values were cropped to $[0,1]$ for rendering.
	}
	\label{fig:mnist_samples}
\end{figure}

\begin{figure}[tb]
	\def\lwidth{8em}
	\def\lheight{7.5em}
	\begin{subfigure}{\lwidth}
		\centering
		\includegraphics[height=\lheight]{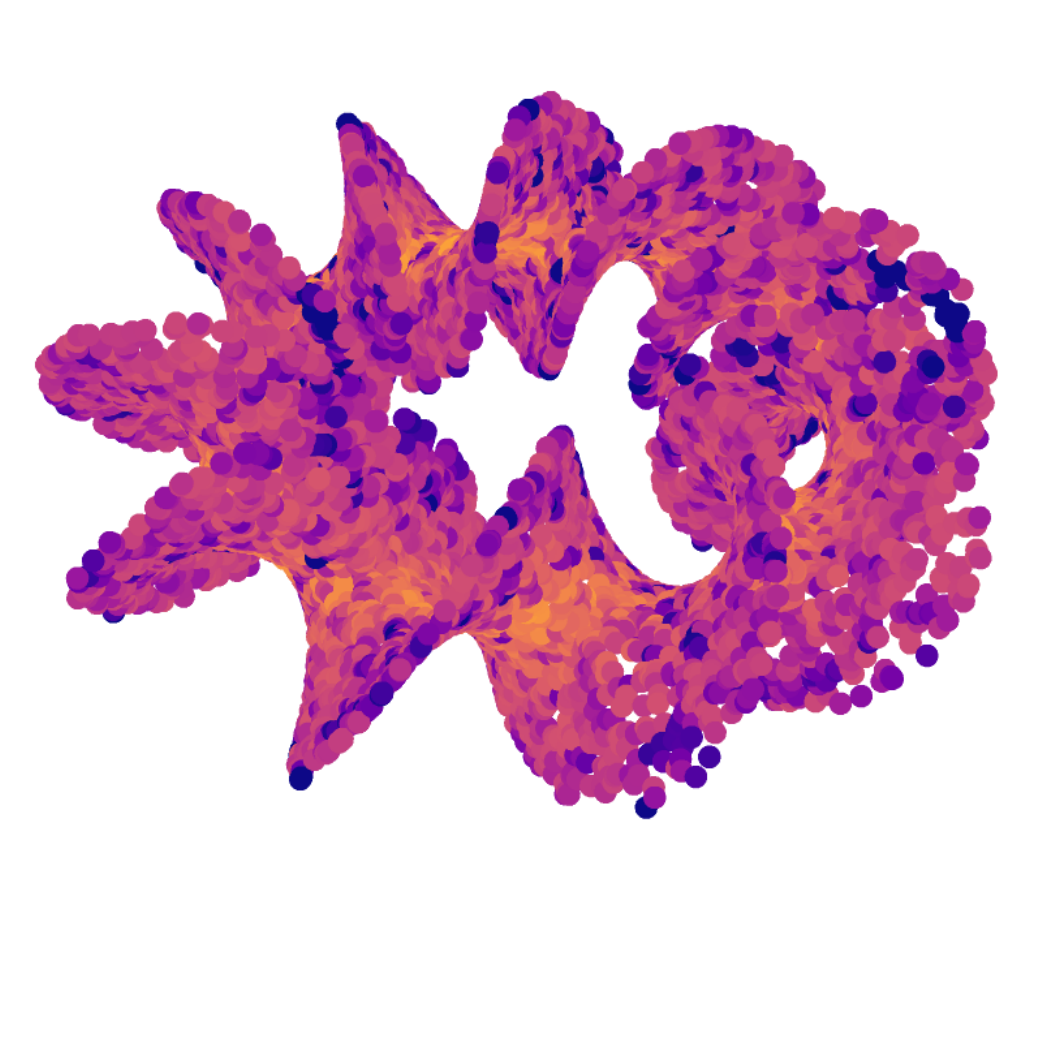}
		\\[-1.5em]
		{\scriptsize without MESS}
	\end{subfigure}
	\begin{subfigure}{10em}
		\centering
		\includegraphics[height=\lheight]{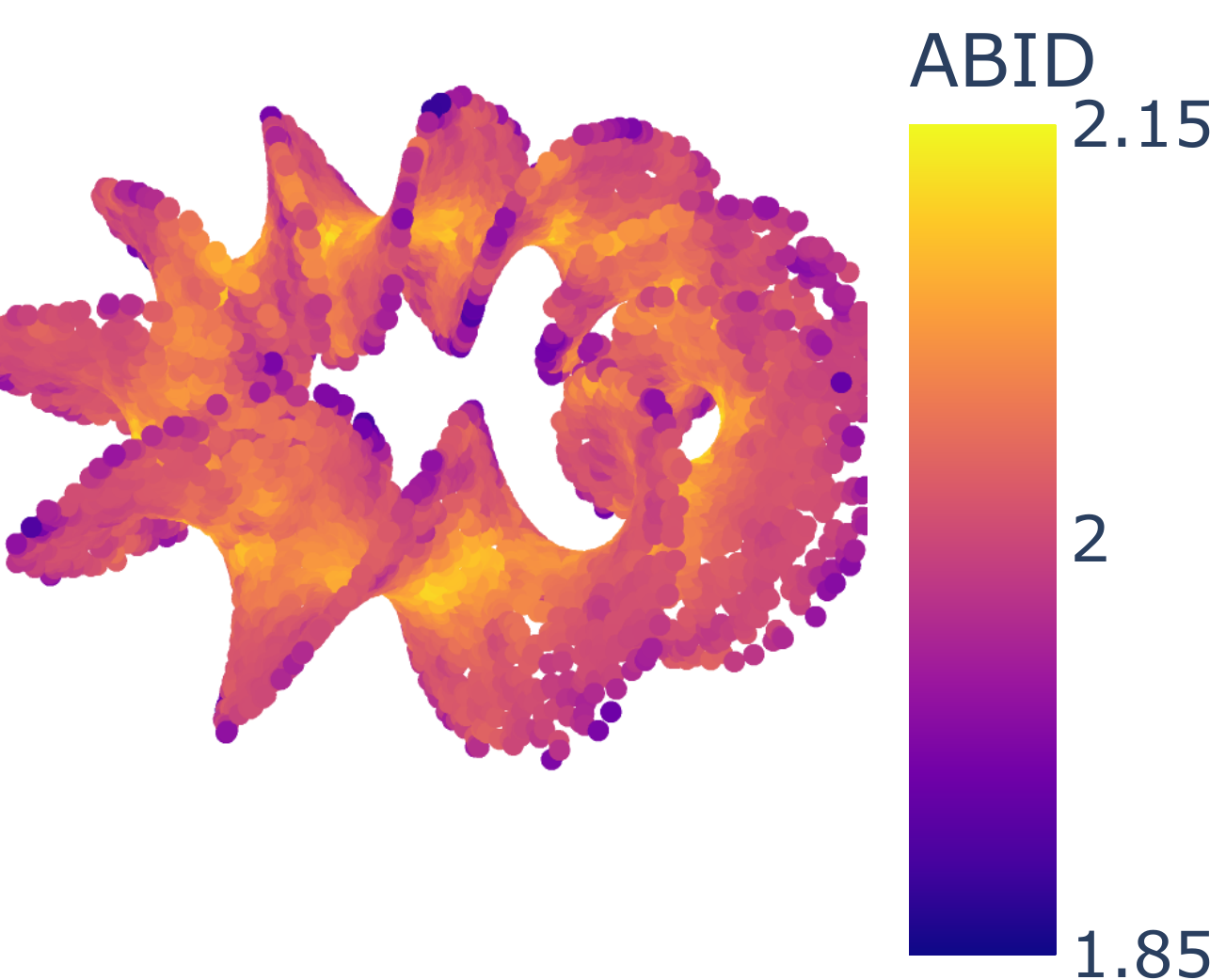}
		\\[-1.5em]
		{\scriptsize with MESS\hspace*{3em}}
	\end{subfigure}
	\begin{subfigure}{9em}
		\includegraphics[height=\lheight]{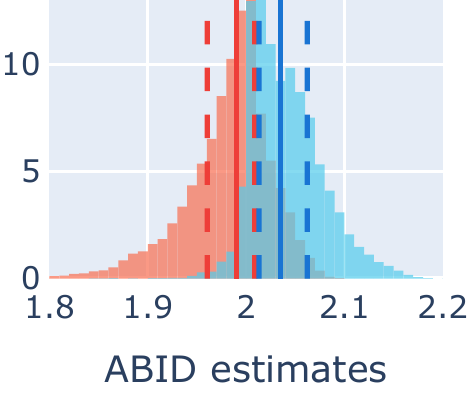}
	\end{subfigure}
	\begin{subfigure}{9em}
		\includegraphics[height=\lheight]{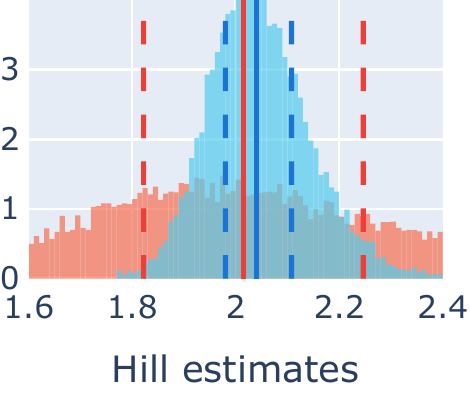}
	\end{subfigure}
	\caption{
		3D plots of the $\mathtt{m11}$ data set colored by ABID estimates without (left) and with (right) supersampling and ID histograms without (orange) and with (blue) supersampling.
		The solid lines are at the median and the dashed lines at the first and third quartile.
		Parameters for these plots are $\mathtt{k_1}{=}\mathtt{k_2}{=}50,\mathtt{k_3}{=}5000,\mathtt{ext}{=}100$ with covariance generation and $C_2+W_3$ correction.
	}
	\label{fig:moebius_scatter}
\end{figure}

We made experiments on multiple synthetic data sets like the \texttt{m1} through \texttt{m13} sets introduced by Rozza et al.~\cite{DBLP:journals/ml/RozzaLCCC12} in part using generators by Hein et al.~\cite{Hein2005IntrinsicDE}, which have been used repeatedly to evaluate ID estimate quality and the toy examples from evaluations on the ABID estimator \cite{DBLP:conf/sisap/ThordsenS20}.
In addition to that, we experimented on a few real data sets like point clouds from 3D scans or MNIST which consists of $28 \times 28$ grayscale images which we interpret as $784$-dimensional vectors.
In our experiments, we could observe qualitative improvement due to the proposed correction.
On MNIST, the corrected samples are less distorted than the raw samples as can be seen in Fig.~\ref{fig:mnist_samples}.
The corrected samples tend to be \enquote{between} the original image and their nearest neighbor in the original set.
Their shape is similar to the original images yet they are not mere linear interpolations which suggests that they lie on the manifold.
Qualitative observations about the correction step like those in Fig.~\ref{fig:pointcloud:swiss_roll} and Fig.~\ref{fig:mnist_samples} were made throughout all humanly visualizable data sets.
The visualizing approach for quality comparison entails a subjective component, which is arguably undesired.
Yet, only arguing ID estimate quality in terms of histograms and summary statistics can be misleading.
Fig.~\ref{fig:moebius_scatter} displays ID estimates without and with supersampling on the $\mathtt{m11}$ data set.
The median of the ABID estimates moves away from the ground truth ($\delta=2$) while the interquartile range barely changes, which would hint at a lower estimate quality.
The 3D plots however show that the ABID estimates are better fitted to the geometry with less local variance.
The better summary statistics without supersampling hide the fact that these estimates are less helpful in understanding the data set.
The Hill estimates are also improved by using supersampling.

As for the generation variants (covariance, Cholesky, Eigendecomposition), we could observe that on the 1000 point subset of MNIST only the samples generated by the covariance-based approach were sufficiently far away from their original samples to be interesting for small $\mathtt{k_1}$.
The Cholesky decomposition-based approach generates samples very close to the original points as expected for $\delta \ll 784$ ($\delta$ of MNIST is suspected to be below 30 \cite{DBLP:conf/kdd/AmsalegCFGHKN15,DBLP:conf/sdm/AmsalegCHKRT19,tr/nii/ChellyHK16,DBLP:conf/sisap/ThordsenS20}).
As for the Eigendecomposition-based approach, either the initial ID estimate ($\approx 4$ as per ABID) was below $\delta$ or the distance bound of one standard deviation is too low.
Both would lead to an increased supersampling density close to the original points.
Using varying initial ID estimates and radius scales, more interesting supersamples can be created for, e.g., an ID of 4 with $5\sigma$ radius, or an ID of 12 with $3\sigma$.
\enquote{Good} radii appear to be both dependent on $\mathtt{k_1}$ and the initial ID estimates, which makes a good choice difficult.
The covariance-based sample generation, therefore, appears to be the overall most promising approach and has been used in all following experiments.

\begin{figure}[tb]
	\def\lwidth{8em}
	\def\lheight{9em}
	\begin{subfigure}{\lwidth}
		\includegraphics[height=\lheight]{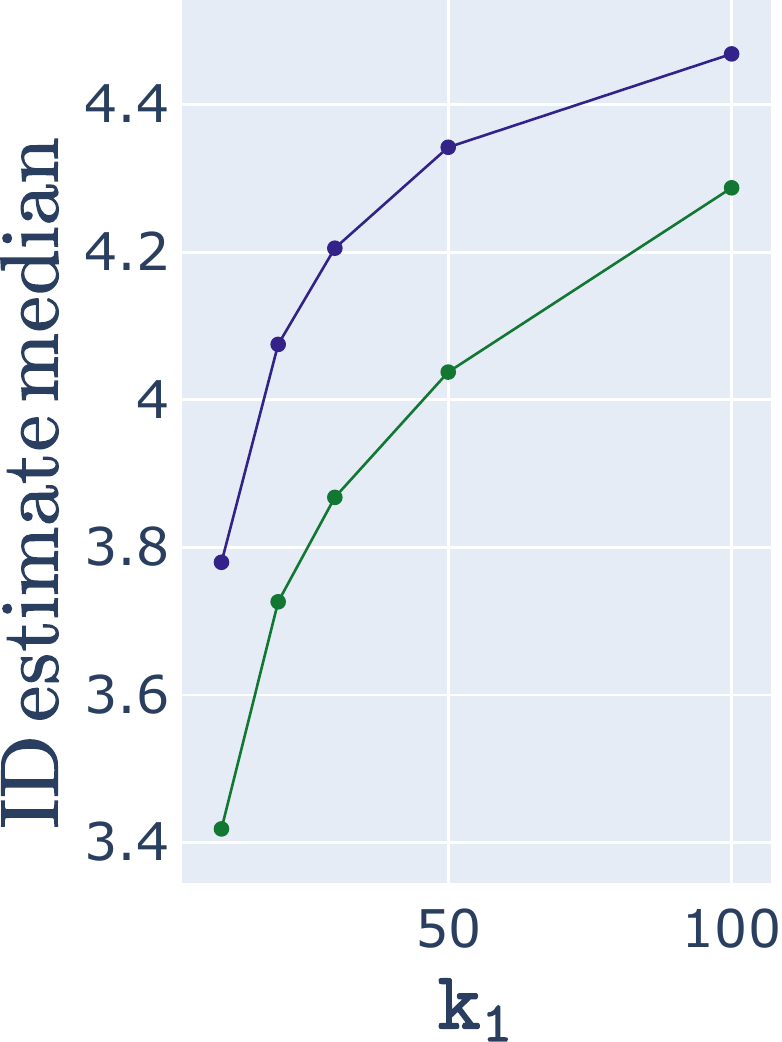}
	\end{subfigure}
	\begin{subfigure}{\lwidth}
		\includegraphics[height=\lheight]{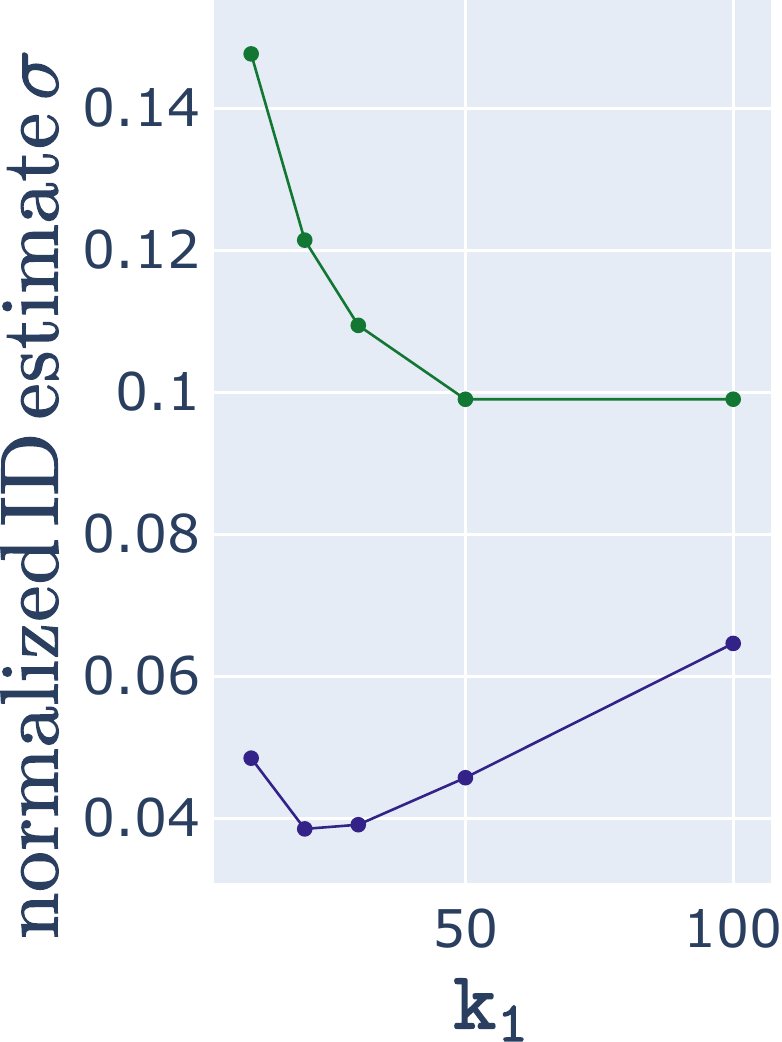}
	\end{subfigure}
	\begin{subfigure}{10em}
		\includegraphics[height=\lheight]{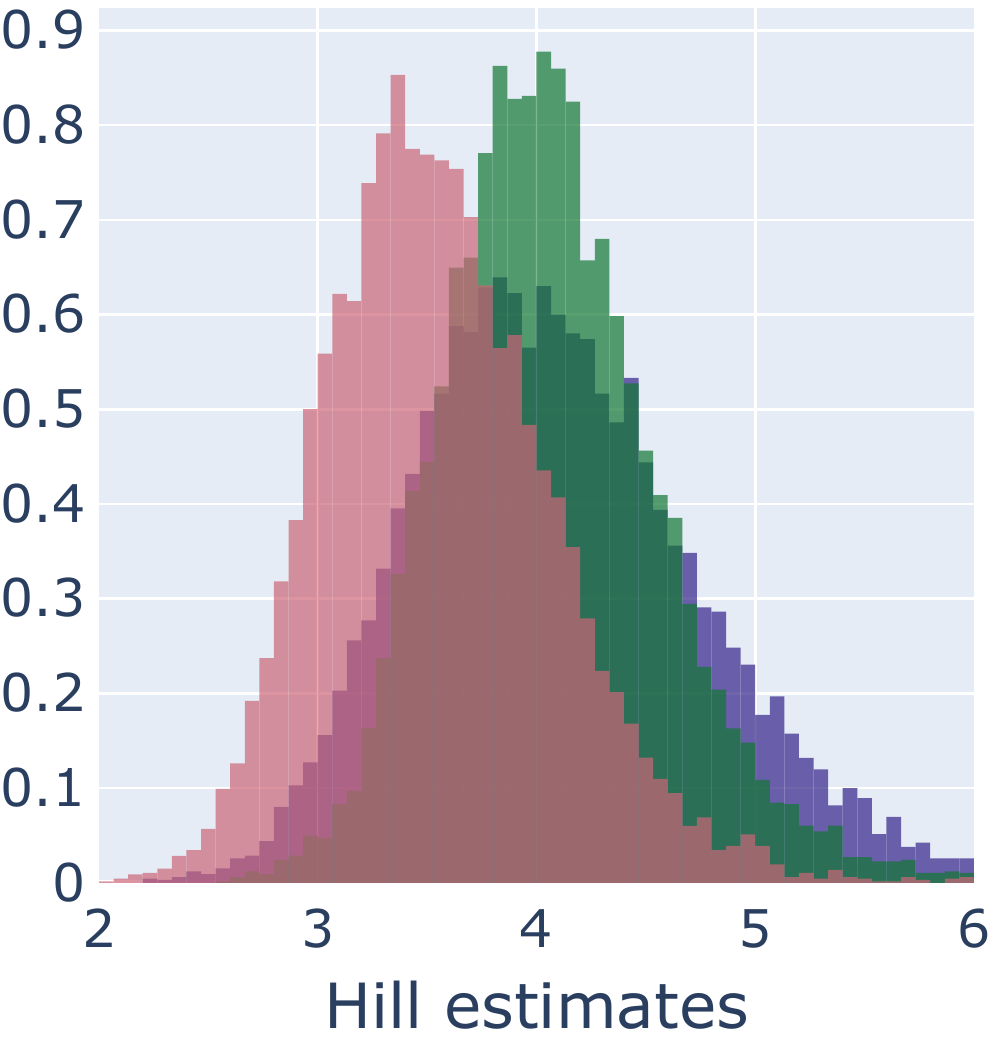}
	\end{subfigure}
	\begin{subfigure}{10em}
		\includegraphics[height=\lheight]{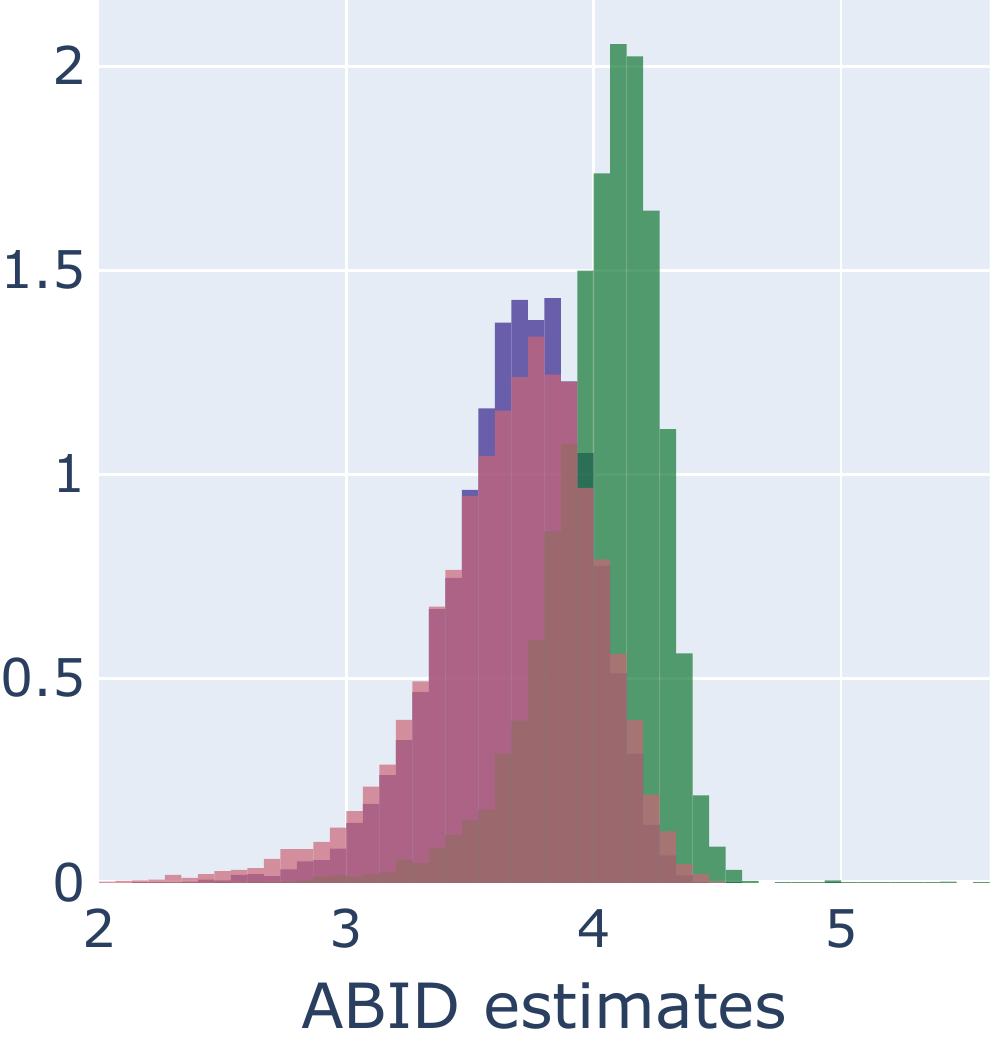}
	\end{subfigure}
	\caption{
		The plots on the left show the impact of varying $\mathtt{k_1}$ for $\mathtt{ext}=75$ on ABID (blue) and Hill (green) estimates on the $\mathtt{m4}$ data set ($d=8,\delta=4$).
		The histograms on the right display the raw estimates (blue) and those obtained with MESS (green) and SMOTE (pink), respectively for the parameters with the lowest local ID standard deviation:
		$\mathtt{k_1}=40$ for Hill and $\mathtt{k_1}=20$ for ABID.
	}
	\label{fig:parameters}
\end{figure}

The correction schemes, consisting of a correction and a weighting rule, give different qualitative results for different estimators.
For the ABID estimator, we observed, that the covariance-based correction $C_1$ can be too aggressive in constraining points onto a sort of \enquote{mean surface}, whereby the original points in noisy or strongly curved data sets tend to lie outside the supersampled set.
In these cases, the estimates become too low, since the ABID estimator gives lower estimates to points on the margin of manifolds.
Using the Cholesky decomposition-based correction $C_2$ does not introduce this effect as strongly.
For the Hill estimator, the correction rules $C_1$ and $C_2$ both give comparable results, since the estimator solely analyzes the expansion rate which is unaffected by points lying on the surface of a manifold.
As for weighting rules, the Mahalanobis-based rules $W_2$ and $W_3$ appear to give the best results whenever the original data set has a high enough sampling density to obtain a good enough approximation of the local gradient via the covariance matrix.
The euclidean $W_1$ rule, which depends less on the original data sampling density, mostly gives similar results whilst generally being slightly inferior.
On extremely sparse data sets like the $\mathtt{m10c}$ data set (10000 points from a 24-dimensional hypercube in 25 dimensions), none of the approaches gave near-ground-truth results, which, however, is also not achieved without supersampling.
For general purposes, the overall best results were achieved using a combination of $C_2$ and $W_3$ with either estimator.
The following paragraphs only consider this combination.

The $\mathtt{ext}$ factor does not appear to have much of an impact on Hill and ABID estimates for $\mathtt{ext} \geq 25$.
Larger $\mathtt{ext}$ values of course yield better results at the cost of additional computation time, as $\mathtt{k_3}$ should be increased with $\mathtt{ext}$.
This observation, however, does not readily generalize to other estimators.
For the ALID estimator, the choice of $\mathtt{ext}$ can have a larger impact even for $\mathtt{ext} > 100$, which largely increases computational cost, as ALID has a computational cost quadratic to the neighborhood size.
Yet, the median ALID estimate still appears to converge for larger $\mathtt{ext}$ values.
For varying $\mathtt{k_1}$ we observed good estimate quality close to or even below $\delta^2$ for $\mathtt{k_1}$.
In these small neighborhoods ID estimators tend to have difficulties getting good estimates \cite{DBLP:conf/sdm/AmsalegCHKRT19,tr/nii/ChellyHK16,DBLP:conf/sisap/ThordsenS20}.
Yet, as smaller neighborhoods give a better approximation of the local manifold structure, lowering the number of neighbors on which we can perform ID estimation, is very interesting.
In most experiments, the estimate quality for increasing $\mathtt{k_1}$ at first improves and afterward deteriorates.
ABID estimates for too low a $\mathtt{k_1}$ tend to be too low.
This can be explained by the neighborhoods being too small to encompass neighbors along each of the orthogonal components of the manifold.
For larger $\mathtt{k_1}$ the ABID estimates can decrease as the original points tend to lie \enquote{on the surface} of the generated manifold in at least one dimension as weaker features of the manifold structure are suppressed.
This is even observable without any candidate correction, albeit at much larger $\mathtt{k_1}$ values.
While that is not necessarily a problem for other applications like classification, it largely affects geometry-based ID estimation.
They can also steadily increase as the entire curved manifold might geometrically appear up to full dimensional.
The Hill estimator appears to give ever-growing median ID estimates for increased $\mathtt{k_1}$ values approaching the ground truth ID with a growing upper tail.
For very large $\mathtt{k_1}$, the Hill estimator can overestimate the ground truth ID with MESS.
The lower end of the Hill estimates, however, appears not to exceed the ground truth ID.
The on-average growth of the estimates can likely be attributed to fluctuations in the sampling density.
The Hill estimate distributions, hence, are highly skewed and require a visual interpretation.
By additionally considering the mean standard deviation of ID estimates of the neighbors divided by the ID estimate of the central point, we can examine the local smoothness of ID estimates.
In our experiments, the $\mathtt{k_1}$ with the smallest mean ID deviation indicated the best ID estimates for both ABID and Hill estimators, except for $\mathtt{m10c}$, where the estimates did not reach 24.
The median Hill estimate of about 20 with MESS was at least closer than the median of about 17 without MESS.
Fig.~\ref{fig:parameters} showcases the impact of $\mathtt{k_1}$ choices and also gives a qualitative comparison to supersampling with SMOTE.

In summary, $\mathtt{ext}$ values of 100 or above are beneficial for ID estimation, although smaller values can suffice for ABID or Hill estimates, and the choice of $\mathtt{k_1}$ largely depends on the data set, where there might be a \enquote{sweet spot} that is large enough to encompass the full complexity of the manifold whilst not exceeding local structures.
Finding that \enquote{sweet spot} can be achieved by analyzing the local deviation of estimates.

\section{Conclusions}\label{sec:conclusion}

In this paper, we introduced a supersampling technique motivated by the geometric form of the embedding function defining the generative mechanism of observed manifolds.
It contrasts preceding methods like those similar to SMOTE by Chawla et al. \cite{Chawla2002SMOTESM} by using the manifold structure.
Yet, it also opposes the method by Bellinger et al. \cite{DBLP:journals/ml/BellingerDJ18} as it does not require an explicit model of the manifold but uses covariances to mimic the Jacobian of the embedding function.
Additional modular correction rules allow compensating for high dimensional noise on the data.
In our experiments, we have shown that the novel approach is capable of generating samples that mimic the manifold structure so that it can even improve ID estimates if sufficiently many supersamples are drawn.
From that, we conclude, that the MESS framework can sample data from the manifold populated by the given data.
It allows to analyze data sets on smaller scales that satisfy the locality assumption of ID estimators.
The improved ID estimates in return support the claim, that the manifold has been properly supersampled, for other applications, like classification.
Additionally, a scan of different $\mathtt{k_1}$ values can help to find reasonable neighborhood sizes using the mean ID deviation.
The major drawback of this technique is the increased time and space requirements due to the increase in the data set and neighborhood sizes.
We did not perform experiments in that regard, but we expect the MESS framework to benefit other machine learning tasks, such as classification, which would be a nearby application beyond ID estimation.

\bibliographystyle{splncs04} %
\bibliography{literature}

\begin{thebibliography}{10}
\providecommand{\url}[1]{\texttt{#1}}
\providecommand{\urlprefix}{URL }
\providecommand{\doi}[1]{https://doi.org/#1}

\bibitem{DBLP:conf/kdd/AmsalegCFGHKN15}
Amsaleg, L., Chelly, O., Furon, T., Girard, S., Houle, M.E., Kawarabayashi, K.,
  Nett, M.: Estimating local intrinsic dimensionality. In: KDD (2015)

\bibitem{DBLP:conf/sdm/AmsalegCHKRT19}
Amsaleg, L., Chelly, O., Houle, M.E., Kawarabayashi, K., Radovanovic, M.,
  Treeratanajaru, W.: Intrinsic dimensionality estimation within tight
  localities. In: {SDM} (2019)

\bibitem{DBLP:journals/ml/BellingerDJ18}
Bellinger, C., Drummond, C., Japkowicz, N.: Manifold-based synthetic
  oversampling with manifold conformance estimation. Mach. Learn.
  \textbf{107}(3) (2018)

\bibitem{Chawla2002SMOTESM}
Chawla, N., Bowyer, K., Hall, L., Kegelmeyer, W.P.: Smote: Synthetic minority
  over-sampling technique. J. Artif. Intell. Res.  \textbf{16},  321--357
  (2002)

\bibitem{tr/nii/ChellyHK16}
Chelly, O., Houle, M.E., Kawarabayashi, K.: Enhanced estimation of local
  intrinsic dimensionality using auxiliary distances. Tech. Rep. NII-2016-007E,
  National Institute of Informatics (2016)

\bibitem{Erba2019IntrinsicDE}
Erba, V., Gherardi, M., Rotondo, P.: Intrinsic dimension estimation for locally
  undersampled data. Scientific Reports  \textbf{9} (2019)

\bibitem{Hein2005IntrinsicDE}
Hein, M., Wrobel, L.: Intrinsic dimensionality estimation of submanifolds in
  euclidean space. In: ICML 2005 (2005)

\bibitem{doi:10.1214/aos/1176343247}
Hill, B.M.: A simple general approach to inference about the tail of a
  distribution. The Annals of Statistics  \textbf{3}(5) (1975)

\bibitem{DBLP:conf/sisap/Houle17}
Houle, M.E.: Local intrinsic dimensionality {I:} an extreme-value-theoretic
  foundation for similarity applications. In: {SISAP} (2017)

\bibitem{DBLP:conf/sisap/Houle17a}
Houle, M.E.: Local intrinsic dimensionality {II:} multivariate analysis and
  distributional support. In: {SISAP} (2017)

\bibitem{DBLP:conf/sisap/Houle20}
Houle, M.E.: Local intrinsic dimensionality {III:} density and similarity. In:
  {SISAP}. Lecture Notes in Computer Science, vol. 12440 (2020)

\bibitem{DBLP:conf/icdm/HouleKN12}
Houle, M.E., Kashima, H., Nett, M.: Generalized expansion dimension. In: {ICDM}
  Workshops (2012)

\bibitem{DBLP:conf/sisap/HouleSZ18}
Houle, M.E., Schubert, E., Zimek, A.: On the correlation between local
  intrinsic dimensionality and outlierness. In: {SISAP} (2018)

\bibitem{Joarder2008OnTD}
Joarder, A., Al-Sabah, W.S., Omar, M.H., Fahd, K.: On the distributions of
  norms of spherical distributions (2008)

\bibitem{DBLP:journals/ml/RozzaLCCC12}
Rozza, A., Lombardi, G., Ceruti, C., Casiraghi, E., Campadelli, P.: Novel high
  intrinsic dimensionality estimators. Mach. Learn.  \textbf{89}(1-2) (2012)

\bibitem{sengupta2003linear}
Sengupta, D., Jammalamadaka, S.R.: Linear models: an integrated approach. World
  Scientific (2003)

\bibitem{DBLP:conf/sisap/ThordsenS20}
Thordsen, E., Schubert, E.: {ABID:} angle based intrinsic dimensionality. In:
  {SISAP} (2020)

\bibitem{wang2006classification}
Wang, J., Xu, M., Wang, H., Zhang, J.: Classification of imbalanced data by
  using the smote algorithm and locally linear embedding. In: Int. Conf. Signal
  Processing. vol.~3 (2006)

\end{thebibliography}
\end{document}